%% file: UGC.tex
\documentclass[10pt,twocolumn,letterpaper]{article}

\usepackage{caption}
\usepackage{iccv}
\usepackage{times}
\usepackage{epsfig}
\usepackage{graphicx}
\usepackage{multirow}
\usepackage{amsmath}
\usepackage{amssymb}
\usepackage{pifont}
\usepackage{algorithm}
\usepackage{algpseudocode}
\usepackage{multicol}
\usepackage[normalem]{ulem}

\usepackage{booktabs} 
\usepackage[accsupp]{axessibility}

\usepackage[table, svgnames, dvipsnames]{xcolor}
\usepackage{amsmath}
\definecolor{darkergreen}{RGB}{19, 182, 53}

\usepackage[pagebackref=true,breaklinks=true,letterpaper=true,colorlinks,bookmarks=false]{hyperref}

\iccvfinalcopy 

\def\iccvPaperID{4426} 

\ificcvfinal\fi
\begin{document}

\title{UGC:  Unified GAN Compression for Efficient Image-to-Image Translation}


\author{Yuxi Ren$^{\star}$ \quad
Jie Wu$^{\star}$$^\dagger$ \quad
Peng Zhang \quad
Manlin Zhang \\
Xuefeng Xiao \quad
Qian He \quad
Rui Wang \quad
Min Zheng \quad
Xin Pan \\
ByteDance Inc \\
\tt\small
\begin{tabular}{@{}l@{}}
\{renyuxi.20190622, wujie.10, zhangpeng.ucas, zhangmanlin.12, xiaoxuefeng.ailab, \\
1988heqian, ruiwang.rw, zhengmin.666, panxin.321\}@bytedance.com
\end{tabular}}


\twocolumn[{%
\renewcommand\twocolumn[1][]{#1}%
\maketitle
\begin{center}
    \vspace{-25pt}
    \includegraphics[width=0.98\textwidth]{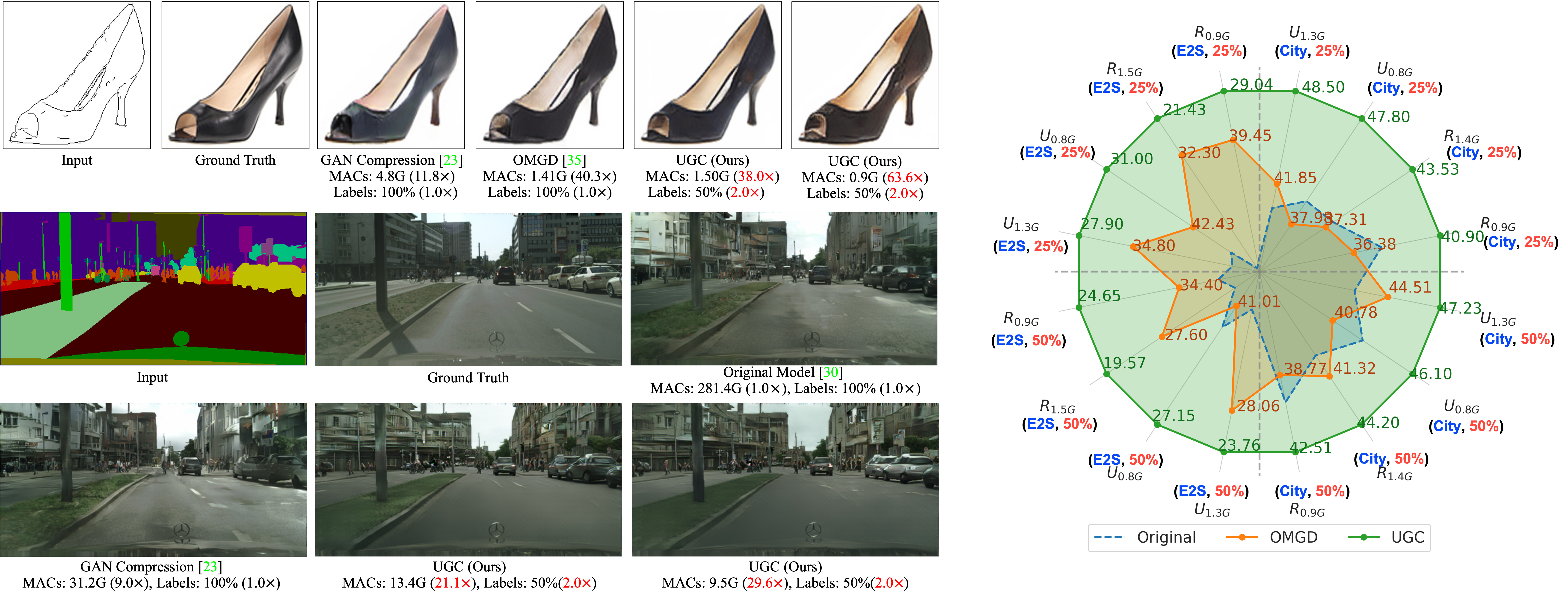}
    \captionof{figure}{
We introduce  Unified GAN Compression (UGC)  for compressing image-to-image translation-based GANs. UGC reduces \textbf{29.6-63.6$\times$} MACs and \textbf{50\%} of data labels while preserving visual fidelity. UGC establishes new state-of-the-art performance across various model constraints (12-67$\times$ MACs), label constraints (\textcolor{red}{10\%}, \textcolor{red}{25\%}, \textcolor{red}{50\%}), datasets (Cityscapes (\textcolor{blue}{City})\cite{Cordts2016TheCD}, Edges2Shoes (\textcolor{blue}{E2S})\cite{Yu2014FineGrainedVC}) and generator architectures (ResNet\cite{resnet, pix2pix}, UNet\cite{unet, pix2pix}). $R_{0.9G}$ denotes the Resnet-style generator with 0.9G MACs. 
\textcolor{magenta}{\href{https://github.com/bytedance/UGC.}{Our code and models are made public at: https://github.com/bytedance/UGC}}
}
    \vspace{3pt}
    \label{fig:head}
\end{center}%
}]

\renewcommand{\thefootnote}{}
\footnotetext{$^\star$Equal contribution. $^\dagger$Corresponding author.}

\ificcvfinal\thispagestyle{empty}\fi

\input{sections/0_Abstract.tex}
\input{sections/1_Intro.tex}

\input{sections/2_RelatedWork.tex}
\input{sections/3_Method.tex}
\input{sections/4_Exp.tex}

\input{sections/5_Conclusion.tex}

{\small
\bibliographystyle{ieee_fullname}
\bibliography{egbib}
}

\end{document}

%% file: sections/0_Abstract.tex
\begin{abstract}
Recent years have witnessed the prevailing progress of Generative Adversarial Networks (GANs) in image-to-image translation.
However, the success of these GAN models hinges on ponderous computational costs and labor-expensive training data. Current efficient GAN learning techniques often fall into two orthogonal aspects: i) model slimming via reduced calculation costs; ii) 
data/label-efficient
learning with fewer training data/labels. 
To combine the best of both worlds, we propose a new learning paradigm, Unified GAN Compression (UGC), with a unified optimization objective to seamlessly prompt the synergy of model-efficient and 
label-efficient learning. 
UGC sets up semi-supervised-driven network architecture search and adaptive online semi-supervised distillation stages sequentially, which formulates a heterogeneous mutual learning scheme to obtain an architecture-flexible, 
label-efficient, and performance-excellent model.
Extensive experiments demonstrate that UGC obtains state-of-the-art lightweight models even with less than \textbf{50\%} labels. UGC that compresses \textbf{40$\times$} MACs can achieve 21.43 FID on edges→shoes with \textbf{25\%} labels, which even outperforms the original model with 100\% labels by 2.75 FID.
\end{abstract}

%% file: sections/1_Intro.tex
\section{Introduction}
\label{sec:intro}

\begin{figure}[t]
  \centering
  \includegraphics[width=0.45\textwidth]{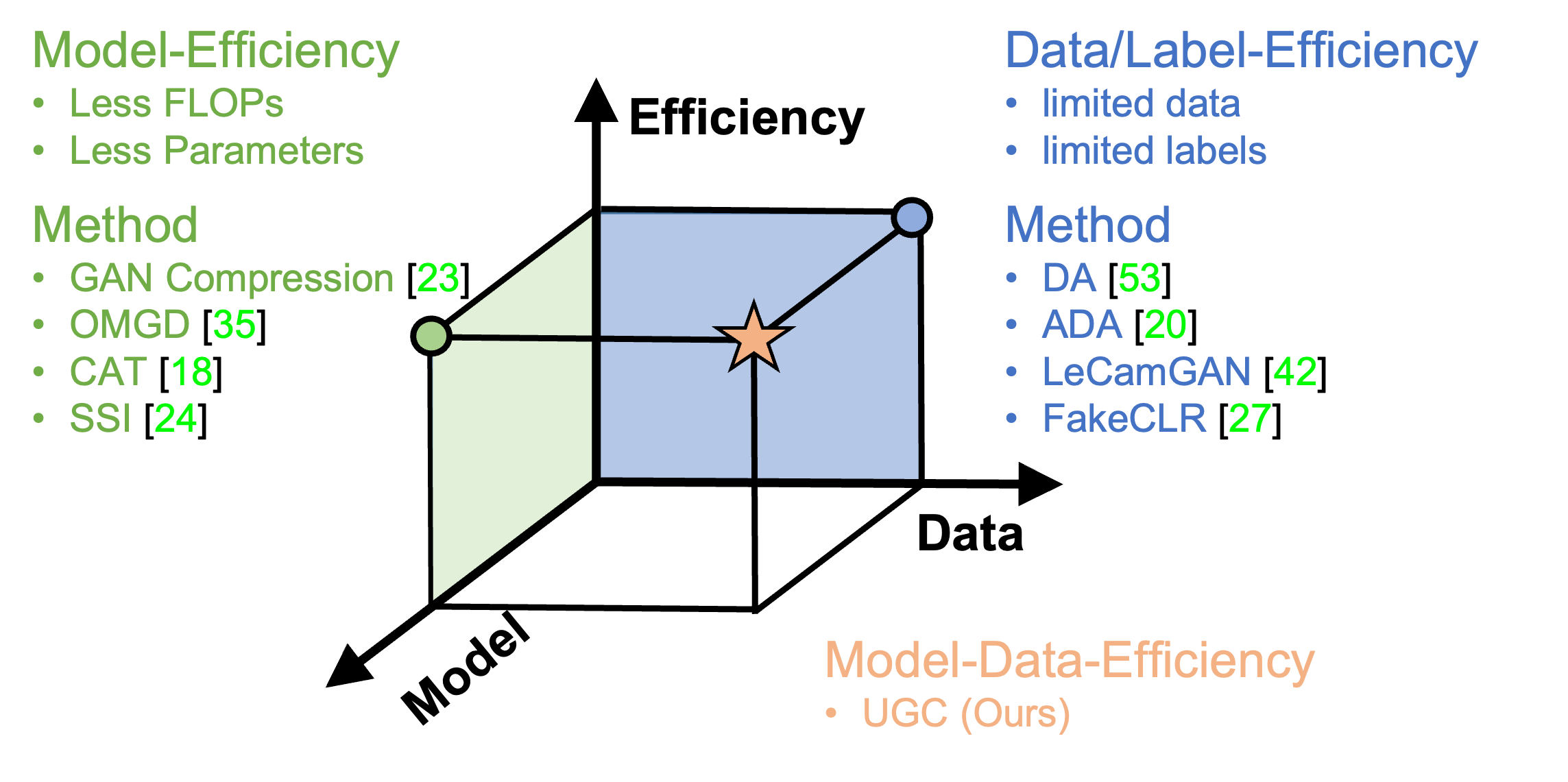}
  \caption{Unified GAN compression in the \textcolor{orange}{model-data-efficiency} space, which consists of \textcolor{LimeGreen}{model-efficient} learning to reduce computational costs, and \textcolor{Periwinkle}{data-efficient} learning with fewer labels.}
  \vspace{-5pt}
    \label{fig:pipeline}
\end{figure}

Recently, Generative Adversarial Networks (GANs)\cite{gan, WGAN, DCGAN, DCGANs, bigGANs, pix2pix, cyclegan}  have achieved prominent results in various visual generative tasks, such as image-to-image translation\cite{pix2pix, wang2018pix2pixHD, GauGAN, cyclegan, StarGAN} and style transfer\cite{StyleGAN, Karras2019stylegan2, stylegan_v, dinh2022hyperinverter}. 
Albeit with varying degrees of progress, most of its recent successes rely on explosive computational complexities or extensive labeled images. 
Training or deploying these excellent GAN models with unwieldy resource demands is arduous, especially in the hardware-constrained\cite{GAN_compression, ren2021online, CAT} or low-data regime\cite{zhao2020differentiable, cao2021remix, li2022fakeclr}.
To alleviate this issue, efficient GAN learning has become a newly-raised and crucial research topic in recent years.
We analyze current corresponding algorithms and classify them into a unified model-data-efficiency space in Figure \ref{fig:pipeline}, which reveals that related techniques can be divided into two orthogonal aspects: model-efficient or data-efficient. 
As shown in Figure \ref{fig:pipeline} left part, model-efficient approaches\cite{GAN_compression, ren2021online, li2021revisiting, CAT, zhang2022wavelet, li2022efficient} typically make efforts to learn slimming GANs to reduce ponderous computational costs and hulking memory usage.
On the other hand, data-efficient learning\cite{cao2021remix, zhao2020differentiable, karras2020training, li2022fakeclr} has recently emerged as a promising approach by leveraging few-shot learning\cite{ojha2021few, zhao2022closer} or regularization techniques\cite{cao2021remix, zhao2020differentiable, karras2020training, tseng2021regularizing}  to optimize GANs with limited training labels, as shown in Figure \ref{fig:pipeline} right part. 

Based on the above observation, we can find that although these two learning paradigms both reside in the common model-data-efficiency space, current
image-to-image translation based GAN compression works only focus on a unilateral optimization objective.
A natural question is: \textit{whether these two orthogonal compression paradigms can be learned uniformly in a collaborative framework?}
Compared with individual learning objectives, a unified solution confronts two additional challenges: 
i) how model and data-efficient learning cooperate in an unified GAN compression framework?
ii) how to overcome the rising challenge of GAN training instability with multiple compression tools?

In many real-world scenarios for image-to-image translation tasks, the labels consistently occupy 50\% of the image input, making it challenging and time-consuming to obtain large amounts of labeled data. In this work, our main focus is to study label efficiency in data-efficient GANs and provide the first attempt to answer the above questions.
We introduce a novel GAN compression objective called Unified GAN Compression (UGC) that seamlessly promotes both model-efficient and label-efficient learning. The proposed UGC optimization framework comprises two stages: a semi-supervised-driven network architecture search and adaptive online semi-supervised distillation.
Each stage integrates both model-based and 
label-centric compression tools, formulating a heterogeneous mutual learning scheme.
In the first stage, semi-supervised learning optimizes sub-networks within the 
search space via distillation loss on unlabeled images, leveraging the weight-sharing mechanism to facilitate the network architecture search (NAS) procedure.
In the second stage,  the discriminator-free student network is optimized in the online distillation setting,  employing the teacher discriminator to guide the 
label-efficient GAN learning procedure.
The contributions of this paper can be summarized in three-fold:

\begin{itemize}
\item ~\underline{\textit{New Insight}}: To the best of our knowledge, we offer the first attempt to design a unified optimization paradigm for GAN compression, which pioneers to advance image-to-image translation based GAN compression into jointly modeling both model-efficient and 
label-efficient GAN learning. We believe that our work can provide inspired insight and suggests a new path forward in efficient GAN learning.
\item ~\underline{\textit{Unified and Pioneering}}: UGC seamlessly integrates three orthogonal compression techniques, i.e., network architecture search, online distillation and semi-supervised learning, to be jointly optimized in a collaborative framework.  On the one hand, semi-supervised learning provides auxiliary supervision signals to facilitate NAS and the distillation procedure. 
On the other hand, NAS and distillation technology promote obtaining an architecture-flexible and performance-excellent model.
\item  ~\underline{\textit{High Effectiveness}}:
Extensive experiments on edges$\rightarrow$shoes \cite{Yu2014FineGrainedVC} and cityscapes \cite{Cordts2016TheCD}) demonstrate that UGC successes to compress  pix2pix \cite{pix2pix} by \textbf{39$\times$} MACs , pix2pixHD by \textbf{17$\times$} and  GauGAN by \textbf{31$\times$} , and achieves state-of-the-art compression performance (see Figure \ref{fig:head}).
Under this extreme compression ratio, UGC can further reduce \textbf{50\%} of labels without losing the visual fidelity of generated images.
Compared with the existing competitive approaches, UGC helps to obtain better image quality with much less computational costs and labels. UGC provides a feasible solution for deploying GAN in the hardware-constrained and 
label-limited regime.

\end{itemize}

%% file: sections/2_RelatedWork.tex
\section{Related Works}
\begin{figure*}[ht]
    \centering
    \includegraphics[width=17cm,height=8cm]{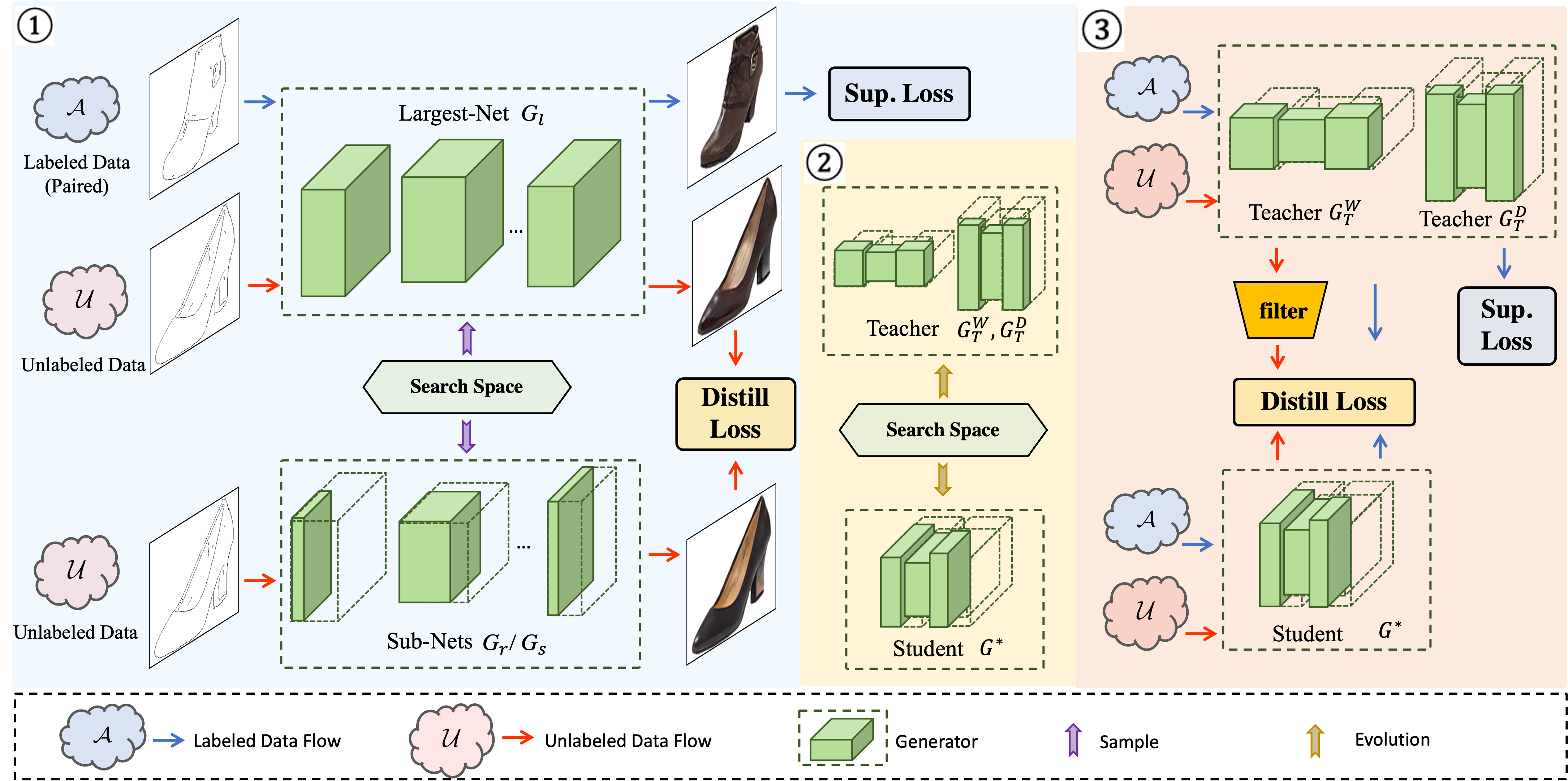}
    \caption{
    The whole pipeline of Unified GAN Compression (UGC) framework. 
    UGC framework contains a two-stage optimization procedure: semi-supervised network architecture search (Step \ding{172}) and adaptive online semi-supervised distillation(Step \ding{173} + Step \ding{174}).
    Step \ding{172}: A largest-net $G_l$ and sub-nets $G_r$, $G_s$ are sampled from the search space and mutually optimized via a weight-sharing mechanism.
    Step \ding{173}: Teacher models $G_T^W$,$G_T^D$ and student models $G^*$ are evolutionary searched from the search space with respect to computational constraints.
    Step  \ding{174}: Student model $G^*$ is optimized with distillation loss via multi-teachers guidance.
    The blue arrow depicts the supervised training process with pair datasets $\mathcal{A}$, while the red counterpart denotes the semi-supervised optimization paradigm for unlabeled data $\mathcal{U}$.
    }
    \vspace{-10pt}
    \label{fig:ugc_framework}
\end{figure*}

\subsection{Image-to-Image Translation GANs}
Generative Adversarial Networks (GANs) \cite{gan} have achieved a remarkable leap on a series of image generation tasks, such as image-to-image translation\cite{cyclegan,pix2pix, CartoonGAN, StarGAN, Tang_2019_CVPR, GauGAN},  image synthesis\cite{DCGAN, Reed2016GenerativeAT, SNGAN, Self-Attention-GAN, bigGANs, imagegeneration} and image stylization\cite{StyleGAN}.
Pix2Pix\cite{pix2pix} proposed to employ paired data for image-to-image translation.
Pix2PixHD\cite{wang2018pix2pixHD} further accomplished the high-resolution photorealistic images generation by coarse-to-fine generator and multi-scale discriminator. GauGAN\cite{park2019SPADE}  boosted the performance of semantic image synthesis via spatially-adaptive normalization. 
These GAN models generate high-fidelity images at the cost of bulky computation and expensive data annotation, which restricts their application in the hardware-constrained or low-data regime. 
Although unpaired image translation \cite{cyclegan, StarGAN} tried to address the label constraints by introducing a weakly-supervised setting, they still suffer from extreme generation instability and quality degradation during model slimming or data volume shrinking.

\subsection{Model-Efficient GAN Learning}
GAN-oriented model compression has been a popular research topic due to the extensive application of GANs in real-scene.
Li \etal \cite{GAN_compression} optimized a "once-for-all" generator and sampled the best-performed architecture according to the target computational constraints. 
Ren \etal \cite{ren2021online} proposed an online multi-granularity distillation to optimize the lightweight models in a discriminator-free way.
Li \etal \cite{li2021revisiting} designed a generator-discriminator cooperative compression scheme to improve the adversarial training stabilization in GAN compression.
Zhang \etal \cite{zhang2022wavelet} selectively distill the high-frequency bands of the generated images to transfer high-quality information effectively.
Li \etal \cite{li2022efficient} innovatively proposed a spatially sparse inference technology to achieve acceleration.
Although these methods achieved a prominent balance between computational cost and image quality, they fail to take data efficiency into consideration. When labeled data from the target domain decrease, even the state-of-the-art GAN compression method\cite{ren2021online} still suffers from noticeable performance degradation, as shown in Figure \ref{fig:cmp_sota}. 

\subsection{Data-Efficient GAN Learning} 
Recently, data-efficient GANs has drawn much attention for their advantage of reducing the manual annotation effort, which mitigate GANs degradation on limited 
data/labels primarily by data augmentation, regularization, and pre-training techniques:
1) Data Augmentation approach \cite{zhao2020differentiable, karras2020training, zhao2020image} focus on relieving the discriminator from overfitting by increasing the training data diversity. 
For example, Zhao \etal \cite{zhao2020differentiable} adopted the differentiable augmentation to stabilize the training process of the discriminator.
2) Regularization methods \cite{yang2021data, cao2021remix, tseng2021regularizing, li2022fakeclr} introduce various priors to improve the generalization of GANs. Tseng \etal \cite{tseng2021regularizing} designed a regularization loss derived from LeCam-divergence to restrain the convergence of the discriminator. Li \etal \cite{li2022fakeclr} designed a contrastive learning based method to enhance performance in a limited-data regime.
3) Pre-training methods \cite{grigoryev2022and, ojha2021few} present transfer-learning solutions for data-efficient GANs.
However, these works are not suitable for the realistic hardware-constrained scenario.
To our knowledge, this paper presents the first 
model-label
bi-dimensional efficient GAN learning paradigm.

%% file: sections/3_Method.tex
\section{Unified GAN Compression}

Unified GAN Compression attempts to accomplish GAN-oriented tasks in the hardware-constrained and 
label-limited regime.
Conditional GANs\cite{pix2pix, wang2018pix2pixHD, GauGAN} aim to learn a mapping function between a source domain $ X=\{x_i\}_{i=1}^{N} $ and a target domain $Y$ . 
In the 
label-efficient
setting, we construct the labeled (paired) training set $ \mathcal{A}= \{x^{\mathcal{A}}_i, y^{\mathcal{A}}_i\}_{i=1}^{M}$ and an unlabeled datasets $\mathcal{U}=\{x^{\mathcal{U}}_j\}_{j=1}^{N-M} $.
For the 
label-efficient
setting, we set the labeled proportion (i.e, $\frac{M}{N}$) to 10\%, 25\% and 50\% to accomplish the GAN-oriented tasks.

This paper offers the first attempt to integrate model-efficient and 
label-efficient
algorithms in the GAN tasks, leveraging network architecture search\cite{bigGANs, wang2021attentivenas}, semi-supervised learning\cite{berthelot2019remixmatch, sohn2020fixmatch, wang2020semi} and distillation algorithm\cite{AGD, ren2021online, zhang2022wavelet} to learn efficient GAN models in a cooperative setting. The whole pipeline of the UGC framework is illustrated in Figure \ref{fig:ugc_framework}, which consists of a two-stage training process.
Specifically, semi-supervised-driven network architecture search and adaptive online semi-supervised distillation are introduced in section \ref{section:SSL NAS} and  \ref{section:SSLOD}, respectively.

\subsection{Semi-Supervised Network Architecture Search}\label{section:SSL NAS}
Network Architecture Search (NAS)\cite{yu2020bignas, wang2021attentivenas} is a widely-used model slimming technology for GAN-oriented compression.
However, 
previous methods\cite{AutoGAN, slimmable_GAN, GAN_compression, CAT} rely heavily on the labeled dataset, thus constraining the model performance in the low-annotation regime.
To alleviate this problem, we introduce Semi-Supervised Learning (SSL) into NAS procedure, which can be regarded as auxiliary supervision to relieve the high dependency on labeled data.
As is shown in Step \ding{172} of Figure \ref{fig:ugc_framework},  we construct a depth-width dynamic supernet as the search space $ \mathcal{S}=\{G_i\} $, $G_i$ is a sub-network sampled from $ \mathcal{S} $. 
We leverage sandwich training rule \cite{yu2020bignas} to sample the largest sub-network ${G_{l}}$, smallest sub-network ${G_{s}}$ and a random sub-networks $ G_{r} $ from $ \mathcal{S} $, then formulates the following optimization function:
\begin{equation}\label{eq:1}
        \mathcal{L}_{stage1}=\mathcal{L}^{\mathcal{A}}_{sup}(G_{l}) + \mathcal{L}^{\mathcal{U}}_{dist}(G_{r}, G_{s})
\end{equation}
The largest sub-network $ G_l $ is optimized on the labeled dataset $ \mathcal{A} $ and other attentively sampled sub-networks are supervised by distillation loss on the unlabeled dataset $ \mathcal{U} $.

\textbf{Supervised Largest Sub-Network Training.} 
We follow the training scheduler in \cite{pix2pix, wang2018pix2pixHD, GauGAN} that employs paired samples from dataset $ \mathcal{A} $ to optimize the largest sub-network $ G_{l} $. 
Take Pix2Pix\cite{pix2pix} as an example, $G_{l}$ is trained to map $x^\mathcal{A}$ to $y^\mathcal{A}$ while its corresponding discriminator $D$ is optimized to distinguish the fake images generated by $G_{l}$ from the real images:
\begin{equation}\label{eq:2}
\begin{split}
    \mathcal{L}_{GAN}(G_{l},D)=&\mathbb{E}_{x^{\mathcal{A}},y^{\mathcal{A}}}[\log D(x^{\mathcal{A}},y^{\mathcal{A}})]\\
                           &+\mathbb{E}_{x^{\mathcal{A}}}[\log (1-D(x^{\mathcal{A}},G_{l}(x^{\mathcal{A}}))].
\end{split}
\end{equation}

Meanwhile, a reconstruction loss is introduced to push the output of $G_{l}$ to be close to the ground truth $y^{\mathcal{A}}$:
\begin{equation}\label{eq:3}
    \mathcal{L}_{Recon}(G_{l})=\mathbb{E}_{x^{\mathcal{A}},y^{\mathcal{A}}}[\parallel{y^{\mathcal{A}}-G_{l}(x^{\mathcal{A}})}\parallel_1].
\end{equation}

To sum up, the supervised loss function $ \mathcal{L}^{\mathcal{A}}$ of $ G_{l} $ is:
\begin{equation}\label{eq:4}
    \begin{split}
        \mathcal{L}^{\mathcal{A}}_{sup}(G_{l})=&\arg\min_{G_l}\max_{D}\mathcal{L}_{GAN}(G_{l},D)\\
                                     &+\lambda_{recon}\mathcal{L}_{Recon}(G_{l})
    \end{split}
    \end{equation}

\textbf{Semi-supervised Sub-Networks Training.} 
To reduce training dependency on target domain images, we randomly sample a middle-sized sub-network $ G_{r} $ and the smallest sub-networks $ G_{s} $\cite{yu2020bignas} from the search space $ \mathcal{S} $. 
Then we conduct an online distillation\cite{ren2021online, li2021revisiting} on $\mathcal{U}$ to capture complementary concepts from unlabeled images.

Specifically, we feed source domain images $x^{\mathcal{U}}$ from $ 4.3 $ to $ G_{l} $ to generate the pseudo paired target domain images $ \hat{y} = G_{l}(x^{\mathcal{U}}) $.
Then semi-supervised learning is conducted via
online distillation loss $ \mathcal{L}_{OD} $, where $ G_l $ guides the optimization direction of sub-networks $ G_{r} $ and $ G_{s} $ step by step. This progressive guidance technique ensures sub-networks are no longer deeply bound with the discriminator, which can train more flexibly and obtain further compression:
\begin{equation}\label{eq:6}
\begin{split}
    \mathcal{L}^{\mathcal{U}}_{dist}(G_{r}, G_{s}) &=
    \mathcal{L}_{OD}(G_{l}(x^{\mathcal{U}}), G_{r}(x^{\mathcal{U}})) \\&+ \mathcal{L}_{OD}(G_{l}(x^{\mathcal{U}}), G_{s}(x^{\mathcal{U}}))
\end{split}
\end{equation}
We utilize the distillation loss in OMGD \cite{ren2021online} as our $ \mathcal{L}_{OD} $, which is composed by Structural Similarity (SSIM) Loss\cite{SSIM}, Perceptual Losses\cite{Perceptual} and Total Variation Loss\cite{TV_Loss}.

In short, $ G_l $ is trained on $ \mathcal{A} $ and supervises sub-networks, while sub-networks absorb knowledge from $ \mathcal{U} $ and further promote $ G_l $ via a weight-sharing mechanism.
This mutual assistant strategy between sub-networks and the largest net promotes the whole super-network in a progressive collaborative way, which \textit{overcomes the first challenge} that facilitates model-data mutual learning in a unified framework.

\subsection{Adaptive Online Semi-Supervised Distillation}
\label{section:SSLOD}
As summarized in \cite{GAN_compression, CAT}, the fine-tuning procedure helps to further boost the target generation capability.
So we follow \cite{ren2021online} to conduct an online multi-teachers distillation to facilitate lightweight GAN learning.
We evolutionary search student and teacher models from $ \mathcal{S} $ in the first stage and conduct an adaptive online multi-teachers distillation fine-tuning scheme to achieve better performance in the second stage.
Online distillation formulates the optimization of the student model in the discriminator-free setting, where the teacher model guides the student model progressively and steadily. 
The adaptive semi-supervised online KD loss consists of a supervised part and a semi-supervised counterpart:
\begin{equation}\label{eq:11}
\small
\mathcal{L}_{stage2}=\mathcal{L}_{sup}^{\mathcal{A}}(G_{T}^{D}) + \mathcal{L}_{sup}^{\mathcal{A}}(G_{T}^{W}) + \mathcal{L}_{dist}^{\mathcal{A}}(G^*) + \mathcal{L}_{dist}^{\mathcal{U}}(G^*)
\end{equation} 
Where $G^*$ is the student model and $G_{T}^{W}, G_{T}^{D}$ are two architectural-complementary teacher models.

\noindent \textbf{Evolutionary Architecture Search.}
As is shown in Step \ding{173} of Figure \ref{fig:ugc_framework}, Given a trained super-net with $ \mathcal{S} $ and a target model constraint, we utilize the evolutionary search algorithm to obtain the best-performed sub-network $ G^* $.
Additionally, we evolutionary search the two architectural-complementary teacher models that have about 20 times more computational costs than $ G^* $ from  $ \mathcal{S} $.
Abbreviated as a deeper network $ G_{T}^{D} $ and a wider network $ G_{T}^{W} $. 
We inherit weights from the super-net to initialize $G_{T}^{D} $, $ G_{T}^{W}$ and $ G^* $, which avoids optimizing the target model from scratch and accelerates network convergence speed.
Furthermore, the broad setting of the super-net allows us to flexibly obtain student and teacher models with diverse constraints, eliminating extra work of manually redesigning and training teacher models.

\noindent \textbf{Supervised Student Model Training.} Specifically, $ G_{T}^{D} $ and $ G_{T}^{W} $ are directly optimized as Eq. \ref{eq:4}, and the corresponding teacher discriminators are inherited from the first stage in section \ref{section:SSL NAS}.
For the student network $ G $, the supervised objective can be formulated as :
\begin{equation}\label{eq:8}
\small
    \begin{split}
        \mathcal{L}_{dist}^{\mathcal{A}}(G^* )&=\mathcal{L}_{OD}(G_{T}^{D}(x^{\mathcal{A}}), G^*(x^{\mathcal{A}})) \\ 
                        &+ \mathcal{L}_{OD}(G_{T}^{W}(x^{\mathcal{A}}), G^*(x^{\mathcal{A}}))
\end{split}
\end{equation}

\noindent  \textbf{Semi-Supervised Student Model Training.}
Obviously, Eq.\ref{eq:8} does not involve the target domain label, so $x^{\mathcal{U}} $ can also participate in this optimization process to further enhance model capacity.
However, the generation ability of teacher networks for unlabeled data can fluctuant since they neither can access the paired ground truth nor has powerful generalization ability as  $ G_{l} $ in \ref{section:SSL NAS}.
Therefore, we design a discriminators-guided adaptive  filtering scheme to improve the pseudo-label quality in the distillation procedure:
\begin{equation}\label{eq:9}
\begin{split}
\small
    \Phi(G_{T}^{D}) = \begin{cases}
        RandInt([0,1]),  & D[G_{T}^{D}(x^{\mathcal{U}})] > D_{EMA} \\
        1, &  D[G_{T}^{D}(x^{\mathcal{U}})] < D_{EMA} 
        \end{cases}
\end{split}
\end{equation}
where $D_{EMA}$ is the exponential moving average (EMA) prediction score of the discriminator for generated fake images and $ RandInt([0,1]) $ means to choose from 0 and 1 randomly.
The semi-supervised objective is calculated with the guidance of an adaptive filter:
\begin{equation}
\label{eq:10}
\footnotesize
\begin{split}
\mathcal{L}_{dist}^{\mathcal{U}}( G^* )&=\Phi(G_{T}^{D})\times \mathcal{L}_{OD}(G_{T}^{D}(x^{\mathcal{U}}), G^*(x^{\mathcal{U}})) \\ 
&+ \Phi(G_{T}^{W}) \times \mathcal{L}_{OD}(G_{T}^{W}(x^{\mathcal{U}}), G^*(x^{\mathcal{U}})).
\end{split}
\end{equation}
The experimental results reveal that this distillation fine-tuning scheme attempts to \textit{address the second challenge} and improves the stability of GAN training under multiple compression technologies. 
We illustrate the complete procedure of Unified GAN Compression in Algorithm \ref{alg:UGC}.

\begin{algorithm}[t]
\footnotesize
    \caption{Unified GAN Compression Scheme.}
    \label{alg:1}
    \begin{algorithmic}[1]
    \State \textbf{Input:} 
    labeled dataset $\mathcal{A}$, unlabeled dataset $\mathcal{U}$, search space $ \mathcal{S}=\{G_i\} $, discriminator $D$, update interval $n$.
    \State \textbf{Output:} trained student generator $G^*$
    \State \textbf{\textit{Stage 1: Semi-Sup Network Architecture Search}}
    \State Initialize search space $\mathcal{S}$
    \For {$steps=1,...,T$}
        \State Sample $G_l$, $G_w$, $G_s$ from $\mathcal{S}$
        \State Sample a mini-batch data: $(x^{\mathcal{A}}, y^{\mathcal{A}})\in \mathcal{A}$, $x^{\mathcal{U}}\in \mathcal{U}$
        \State Calculate $\mathcal{L}^{\mathcal{A}}_{sup}(G_{l}, D)$ by Eq.\ref{eq:4}.
        \State Calculate $\mathcal{L}^{\mathcal{U}}_{dist}(G_{w}, G_{s})$ by Eq.\ref{eq:6}
        \State Update parameters of $G_l, G_s, G_w, D$.
    \EndFor
      
    \State \textbf{\textit{Stage 2: Adaptive Online Semi-Sup Distillation}}
    \State \textit{Evolutionary Architecture Search}: Search $G^*$, $G_{T}^{W}$, $G_{T}^{D}$ from $\mathcal{S}$  
    \State Initialize $G^*$, $G_{T}^{W}$, $G_{T}^{D}$, $D$ from \textit{Stage 1}.
    \For {$steps=1,...,T$}
        \State Sample a mini-batch data: $(x^{\mathcal{A}}, y^{\mathcal{A}})\in \mathcal{A}$, $x^{\mathcal{U}}\in \mathcal{U}$
        \If{$steps$  $ \% $  $ n = 0$}
            \State Calculate $\mathcal{L}^{\mathcal{A}}_{sup}(G_{T}^{W}, D)$ by Eq.\ref{eq:4}.
            \State Calculate $\mathcal{L}^{\mathcal{A}}_{sup}(G_{T}^{D}, D)$ by Eq.\ref{eq:4}.
            \State Update parameters of $G_{T}^{W}$, $G_{T}^{D}$, $D$.
        \EndIf
        \State Calculate $\mathcal{L}_{dist}^{\mathcal{A}}(G^*)$ by Eq.\ref{eq:8}
        \State Calculate $\mathcal{L}_{dist}^{\mathcal{U}}(G^*)$ by Eq.\ref{eq:10}
        \State Update parameters of $G^*$
    \EndFor
    \end{algorithmic}
\label{alg:UGC} 
\end{algorithm}

%% file: sections/4_Exp.tex
\section{Experiments} \label{sec:experiments}
\subsection{Experimental Settings}
\input{./tables/cmp_sota_city}
\input{./tables/cmp_with_sota_pix2pix}
We follow models, datasets and evaluation metrics used in previous works \cite{AGD, GAN_compression,DMAD,co_evolution} for a fair comparison.

\begin{figure*}[t]
  \centering
  \includegraphics[width=\linewidth]{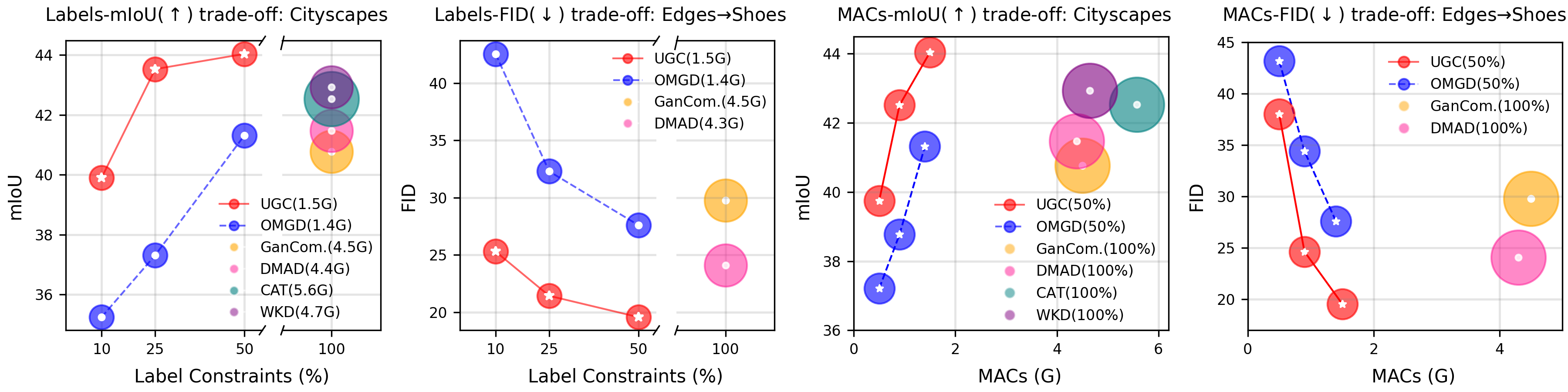}
  \vspace{-15pt}
  \caption{MACs-/Labels-Performance trade-off between UGC and state-of-the-art approaches including GAN compression\cite{GAN_compression}, DMAD\cite{DMAD}, CAT\cite{CAT}, OMGD\cite{ren2021online} and WKD\cite{zhang2022wavelet}.
The first two figures depict the trade-off in different label constraints. A larger circle denotes that the model has greater MACs. The last two figures illustrate the results under different model constraints, where the circle size represents the usage of the labels. UGC significantly outperforms these methods with much less computational cost and labeled data.}
  \vspace{-10pt}
    \label{fig:cmp_sota}
\end{figure*}

\textbf{Models.} 
Pix2Pix \cite{pix2pix}  is the widely-used model in GAN-oriented model compression \cite{GAN_compression, ren2021online, li2021revisiting, CAT, zhang2022wavelet}. We follow \cite{GAN_compression, ren2021online} that adopts the UNet\cite{unet} and the ResNet style\cite{resnet} generator to compress the Pix2Pix model.
Pix2PixHD \cite{wang2018pix2pixHD}  is an image-to-image translation approach advanced in Pix2Pix for higher-resolution images, where the ResNet-style generator with more parameters is adopted.
GauGAN \cite{park2019SPADE}  is a state-of-the-art semantic image synthesis algorithm that employs spatially adaptive (DE)normalization.

\textbf{Datasets and Evaluation Metrics.} 
We conduct the experiments on cityscapes \cite{Cordts2016TheCD} and edges$\rightarrow$shoes \cite{Yu2014FineGrainedVC} dataset.
In the 
label-efficient training procedure, we randomly sample 10\%, 25\%, 50\% data from the training set as the labeled data and treat others as the unlabeled counterpart.
We follow \cite{GAN_compression, ren2021online} to measure mIoU(mean Intersection over Union) on cityscapes and FID\cite{FID}(Frechet Inception Distance) on edges$\rightarrow$shoes to evaluate the quality of the generated image.
Higher mIoU implies the generated images are more realistic. FID compares the distribution of generated images with real images, and a smaller FID means the performance is more convincing.
We use MACs and Parameters to measure the computational costs of the model. 

\textbf{Implementation details.} 
We set the initial learning rate as 0.0002 and linearly decay to zero in all experiments. $\lambda_{SSIM},\lambda_{feature},\lambda_{style},\lambda_{TV}$ in $ \mathcal{L}_{OD} $ are fixed as $1e1, 1e4, 1e1, 1e-5$ respectively.
To construct the search space $ \mathcal{S}$, we first insert 3 resnet blocks after every upsample/downsample layer into the original model.
The sub-networks sampling procedure dynamically selects layer channel number c=[8, 16, ...,64], step=8;  and corresponding resnet blocks d=[0, 1, 2].

\subsection{Comparison with state-of-the-art methods}
In this section, we compare UGC with several state-of-the-art GAN compression approaches and the original models under different label constraints. We reimplemented all competitors under their official codebase and maintained their training settings to conduct the experiments under 10\%, 25\%, and 50\% partition protocols.
As shown in Table \ref{table:cmp_sota_city}, Table \ref{table:result1} and Figure \ref{fig:cmp_sota},  we report the comparison results on edges$\rightarrow$shoes \cite{Yu2014FineGrainedVC} and cityscapes \cite{Cordts2016TheCD} dataset.

\subsubsection{Label Efficiency}
From Table \ref{table:cmp_sota_city} and \ref{table:result1}, we have the following observations:

\noindent \textbf{Overall Performance.} UGC shows consistent significant advantages under 10\%, 25\% and 50\% label constraints, which reveals that UGC help to capture additional data characteristic to facilitate model compression. Compared with a similar MACs model (1.4G$\sim$1.5G) from OMGD, UGC  decreases the FID by \textbf{10.84} on edges$\rightarrow$shoes with \textbf{25\%} labels. And this result even outperforms the original model with \textbf{full} dataset by a large margin (24.18 vs. 21.43). 

\noindent \textbf{Model Type.} UGC generalizes well on different conditional paired image-to-image GANs. 
The UGC-based Pix2PixHD model outperforms the original model measures by mIoU, which increases from 55.50 to 57.28 with 50\% labels.
GauGAN also shows promising results, UGC obtains a \textbf{10.9\%} mIoU improvement on cityscapes under 25\% labels. UGC can surpass the original model trained on 100\% dataset in 50\% label protocols.

\noindent \textbf{Generator Architecture.} UGC has strong robustness for diverse generator architecture.
On cityscapes, UGC with ResNet style generator (dubbed as UGC(R)) improves the mIoU by \textbf{13.2\%, 16.7\%, 7\%} on 10\%, 25\% and 50\% labeled settings respectively.
 UGC with UNet style generator(dubbed as UGC(U)) with \textbf{50\%} labels achieves better performance than OMGD(U) with \textbf{100\%} labeled protocols, declining FID from 25.00 to 23.25 on edges$\rightarrow$shoes.

\begin{figure*}[t]
  \centering
  \includegraphics[width=\linewidth]{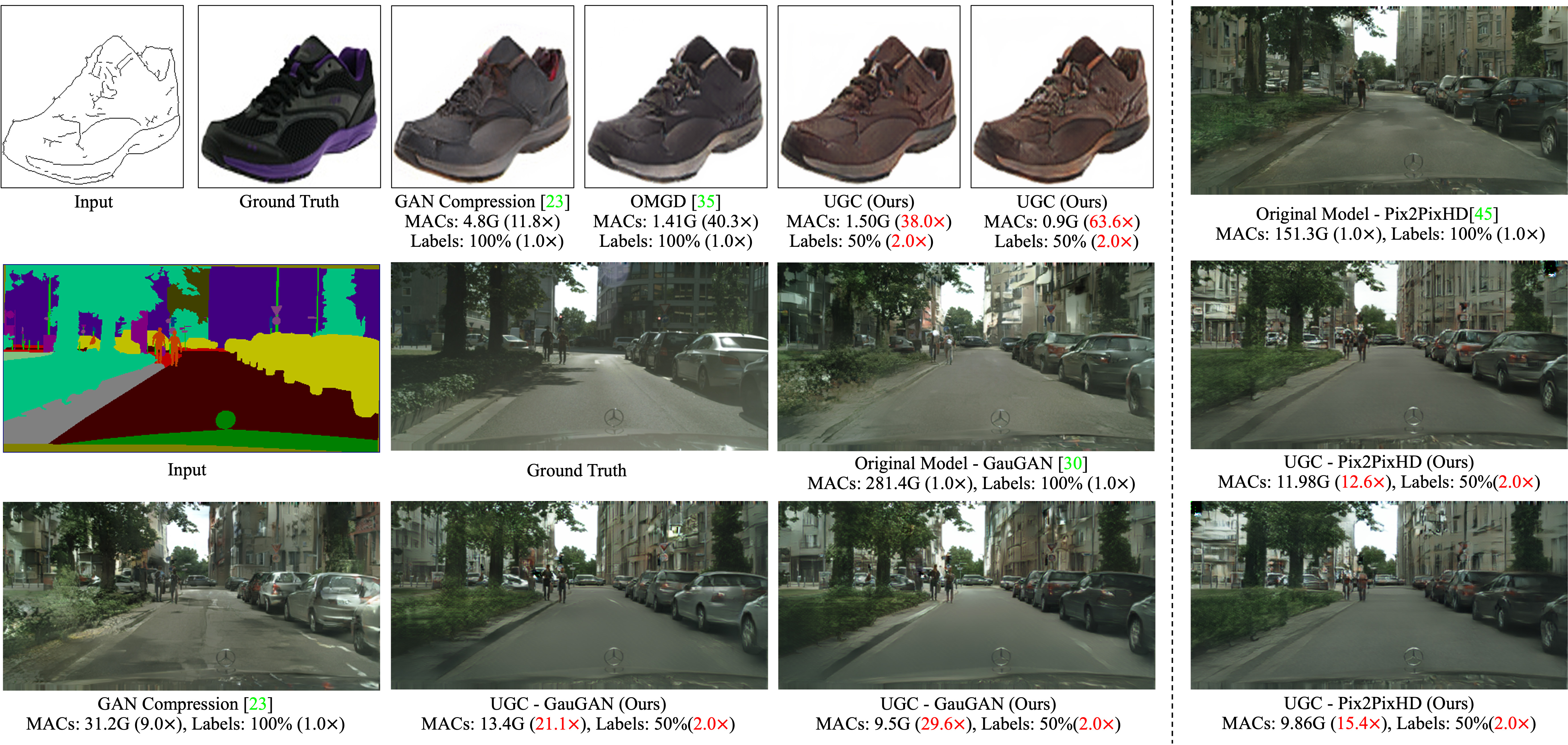}
  \caption{Qualitative Results. UGC reduces MACs by \textbf{38$\times$} and \textbf{29.6$\times$} for Pix2Pix and GauGAN with 50\% labels, respectively.}
    \label{fig:quality_res}
\end{figure*}

\subsubsection{Model Efficiency} 
Essentially, UGC manages to obtain lightweight models in a label-efficient way. We compare the performance from the model-constraint perspective in this section.
Compared to the original model in Pix2Pix, UGC(R) is compressed about \textbf{39$\times$} of MACs and \textbf{75$\times$} of parameters, UGC(U) is compressed by \textbf{15$\times$} of MACs and \textbf{27$\times$} of parameters.
Nevertheless, even under this extreme compression ratio, UGC(R) and UGC(U) surpass the original model by a large margin.
UGC achieves better performance than OMGD on Pix2Pix models with similar compression ratios. Compared to OMGD, UGC(R) and UGC(U) decrease the FID for \textbf{24.4\%} and \textbf{7\%} respectively on edges$\rightarrow$shoes using 50\% labels.
On GauGAN and Pix2PixHD, UGC achieves a similar mIoU with the original model while reducing the computation by \textbf{23$\times$} and half of the labels. 
Conclusively, UGC coordinates model compression well with data compression, achieving state-of-the-art light-weighted models even with less than 50\% labels. 

\input{./tables/ablation_1}
\input{./tables/ablation_2}

\subsection{Ablation Study} 
In this section, we conduct ablation studies to investigate the effectiveness of each essential component of UGC. Experiments are carried out on the UNet style generator of the Pix2Pix model on Cityscapes and edges$\rightarrow$shoes.

\noindent \textbf{Analysis of SSL in NAS.} Firstly, we construct a variant (abbreviated as "Ours w/o NAS-SSL") that trains the super-network in section \ref{section:SSL NAS} without unlabeled dataset $\mathcal{U}$. 
Experiments in Table \ref{table:data efficiency} indicate that the absence of unlabeled images leads to a noticeable drop in performance. For example, "Ours w/o NAS-SSL" increases the FID from 34.8 to 38.16 compared to our approach on the 10\% labeled edges$\rightarrow$shoes dataset.

\noindent \textbf{Analysis of SSL in Finetune.}
Secondly, we design a variant "Ours w/o Finetune-SSL" that removes the unlabeled data usage during the online semi-supervised distillation procedure in section \ref{section:SSLOD}. 
As summarized in Table \ref{table:data efficiency}, without SSL in the finetuning process, the generator declines 2.9\% and 7.9\%  under the 25\% labeled dataset setting on cityscapes and edges$\rightarrow$shoes, respectively. 
The variant "Ours w/o ALL-SSL" without $\mathcal{U}$ in both NAS and finetune process performs worst on all datasets. 

\noindent \textbf{Analysis of NAS.} 
To delve deep into the significance of NAS process, we additionally train a fixed-architecture model with similar MACs on $\mathcal{A}$ and $\mathcal{U}$. As shown in Table \ref{table:model efficiency}, NAS-based UGC obtains 8.6\% and 19.8\% performance improvement under 25\% label setting in 1.2G MACs. The experiment results demonstrate that our proposed NAS technique effectively boosts the model performance in GANs.

\subsection{Hardware Acceleration}
\input{./tables/speed}
We report the inference latency results on the GPU and CPU of Mi 10 using TNN toolkits$\footnote{TNN: https://github.com/Tencent/TNN}$.
As shown in Table \ref{table:speed}, UGC obtains significant inference acceleration on mobile phones. 
For Pix2Pix, the generator compressed by UGC surpasses \textbf{40FPS} inference speed, achieving the demand for interactive applications on mobile phones.  UGC attempts to reduce latency from 2366.7ms to 275.6ms for GauGAN, with a nearly \textbf{90\%} latency decline and outperforms \cite{GAN_compression} by a large margin.
It demonstrates that UGC provides a feasible solution for real-time image translation on edge devices.

\subsection{Qualitative Results}
We depict the visualization results of UGC and the state-of-the-art GAN compression methods in Figure \ref{fig:quality_res}.
Pix2Pix compressed by UGC shows advantageous texture and glossy reconstruction ability on edges$\rightarrow$shoes compared to other methods. UGC on GauGAN and Pix2PixHD are excellent at synthesizing the complex details of buildings and vehicles, achieving the best generation of natural street scenes.

%% file: tables/cmp_sota_city.tex
\begin{table*}[ht]
    \centering
    \footnotesize
    \caption{The performance comparison with state-of-the-art methods on Cityscapes Dataset.}
    \vspace{5pt}
    \renewcommand{\arraystretch}{1.2}
    \begin{tabular}{c|c|c|m{2.2cm}<{\centering}|m{1.7cm}<{\centering}|m{1.7cm}<{\centering}|m{1.6cm}<{\centering}|m{1.6cm}<{\centering}|m{0.8cm}<{\centering}}
    \toprule
    \multirow{2}{*}{Model} &\multirow{2}{*}{Arch.} & \multirow{2}{*}{Method} & \multicolumn{2}{c|}{\textbf{\textit{\textcolor{LimeGreen}{Model Constraints}}}} & \multicolumn{4}{c}{\textbf{\textit{\textcolor{Periwinkle}{Label Constraints}}} @ mIoU ($\uparrow$)}\\
    \cline{4-9}
     & & & MACs & \#Params & 10\% & 25\% & 50\% & 100\% \\
    \toprule
    \multirow{11}{*}{Pix2Pix~\cite{pix2pix}} 
    & \multirow{6}{*}{ResNet} 
    & Original & 56.8G & 11.3M & 34.26 & 37.83 & 40.09 & 44.32 \\
    & & GAN Com. \cite{GAN_compression} & 4G$\sim$5G (13$\times$) & 0.7M (16$\times$) & 36.27 & 38.69 & 41.43 & 40.77 \\
    & & OMGD \cite{ren2021online} & 0.87G (65$\times$) & 0.08M (141$\times$) & 32.90 & 36.38 & 38.77 & 42.15 \\
    & &  UGC &0.8G$\sim$0.9G (\textcolor{red}{67$\times$}) & 0.10M (\textcolor{red}{113$\times$}) & \textbf{37.78 (+3.52)} & \textbf{40.90 (+3.07)}  & \textbf{42.51 (+2.42)} & --- \\
    & & OMGD \cite{ren2021online} & 1.41G (40$\times$) & 0.14M (81$\times$) & 35.24 & 37.31 & 41.32 & 45.21 \\
    & & UGC & 1.4G$\sim$1.5G (\textcolor{red}{39$\times$}) & 0.15M (\textcolor{red}{75$\times$}) & \textbf{39.90 (+5.64)} & \textbf{43.53 (+5.70)} & \textbf{44.20 (+4.11)} & --- \\
    \cline{2-9}
    & \multirow{5}{*}{UNet} 
    & Original & 18.6G & 54.5M & 33.36 & 40.10 & 42.78 & 42.71 \\
    & & OMGD \cite{ren2021online} & 0.71G (26$\times$) & 1.9M (29$\times$) & 34.99 & 37.98 & 40.78 & 45.52 \\
    & & UGC & 0.8G$\sim$0.9G (\textcolor{red}{22$\times$}) & 1.5M (\textcolor{red}{36$\times$}) & \textbf{44.83 (+11.47)} & \textbf{47.80 (+7.70)} & \textbf{46.10 (+3.32)} & --- \\
    & & OMGD \cite{ren2021online} & 1.22G (15$\times$) & 3.4M (16$\times$) & 35.47 & 41.85 & 44.51 & 48.91 \\
    & & UGC & 1.2G$\sim$1.3G (\textcolor{red}{15$\times$}) & 2.0M (\textcolor{red}{27$\times$}) & \textbf{44.46 (+11.10)} & \textbf{48.50 (+8.40)} & \textbf{47.23 (+4.45)} & --- \\
    \toprule
    \multirow{3}{*}{Pix2PixHD~\cite{wang2018pix2pixHD}}
    & \multirow{3}{*}{ResNet} 
    & Original & 151.3G & 182.6M & 45.66 & 49.66 & 55.50 & 58.30 \\
    & & UGC
    & 8G $\sim$ 10G (\textcolor{red}{17$\times$}) & 5.0M (\textcolor{red}{37$\times$})  & 46.34 (+0.68) & 52.40 (+2.74) & 56.65 (+1.15) & --- \\
    & & UGC & 12G $\sim$ 14G (\textcolor{red}{12$\times$}) & 7.0M (\textcolor{red}{26$\times$})  & \textbf{47.22 (+1.56)}  & \textbf{54.42 (+4.76)} & \textbf{57.28 (+1.78)} & --- \\
    \toprule
    \multirow{4}{*}{GauGAN~\cite{GauGAN}}
    & \multirow{4}{*}{ResNet} 
    & Original & 281.4G & 93.1M & 46.30 & 55.54 & 58.24 & 62.30 \\
    & & GAN Com. \cite{GAN_compression} & 30G $\sim$ 32G (9$\times$) & 20M (5$\times$) & 48.93 & 55.21 & 58.19 & 61.22 \\ 
    & & UGC & 8G $\sim$ 10G (\textcolor{red}{31$\times$}) & 3.0M (\textcolor{red}{31$\times$}) & 54.85 (+8.55) & 60.94 (+5.40) & 62.47 (+4.23) & --- \\
    & & UGC & 12G $\sim$ 14G (\textcolor{red}{22$\times$}) & 5.0M (\textcolor{red}{19$\times$}) & \textbf{56.50 (+10.20)}  & \textbf{61.53 (+5.99)} & \textbf{62.73 (+4.49)}& --- \\
    \bottomrule
    \end{tabular}
    \renewcommand{\arraystretch}{1.}
    \label{table:cmp_sota_city} 
\end{table*}

%% file: tables/cmp_with_sota_pix2pix.tex
\begin{table*}[ht]
    \centering
    \footnotesize
    \caption{The performance comparison with state-of-the-art methods in Pix2Pix model on Edges$\rightarrow$Shoes Dataset.}
    \vspace{5pt}
    \renewcommand{\arraystretch}{1.2}
    \begin{tabular}{c|m{2.2cm}<{\centering}|m{2.4cm}<{\centering}|m{2.4cm}<{\centering}|m{1.8cm}<{\centering}|m{1.8cm}<{\centering}|m{1.8cm}<{\centering}|m{0.9cm}<{\centering}}
    \toprule
    \multirow{2}{*}{Arch.} & \multirow{2}{*}{Method} & \multicolumn{2}{c|}{\textbf{\textit{\textcolor{LimeGreen}{Model Constraints}}}} & \multicolumn{4}{c}{ \textbf{\textit{\textcolor{Periwinkle}{Label Constraints}}} @ FID ($\downarrow$)}\\
    \cline{3-8}
    & & MACs & \#Params & 10\% & 25\% & 50\% & 100\% \\
    \toprule
    \multirow{6}{*}{ResNet} 
    & Original \cite{pix2pix} & 56.8G & 11.3M & 89.78 & 65.80 & 38.18 & 24.18 \\
    & GAN Com. \cite{GAN_compression} & 4G$\sim$5G (13$\times$) & 0.7M (16$\times$) & 59.13 & 34.82 & 29.78 & 26.60 \\
    & OMGD \cite{ren2021online} & 0.87G (65$\times$) & 0.08M (141$\times$) & 44.96 & 39.45 & 34.40 & 34.48\\
    & UGC & 0.8G$\sim$0.9G (\textcolor{red}{67$\times$}) & 0.10M (\textcolor{red}{113$\times$}) & \textbf{33.00} \textbf{(-56.78)} & \textbf{29.04} \textbf{(-36.76)} & \textbf{24.65} \textbf{(-13.53)} & --- \\
    & OMGD \cite{ren2021online} & 1.41G (40$\times$)  & 0.14M (81$\times$)& 42.56 & 32.30 & 27.60 & 25.88 \\
    & UGC & 1.4G$\sim$1.5G (\textcolor{red}{39$\times$}) & 0.15M (\textcolor{red}{75$\times$}) & \textbf{25.30} \textbf{(-64.48)} & \textbf{21.43} \textbf{(-44.37)} & \textbf{19.57} \textbf{(-18.61)} & ---\\
    \toprule
    \multirow{5}{*}{UNet} 
    & Original \cite{pix2pix} & 18.6G & 54.5M & 86.42 & 47.46 & 38.51 & 34.31 \\
    & OMGD \cite{ren2021online} & 0.71G (26$\times$) & 1.9M (29$\times$) & 87.20 & 42.43 & 41.01 & 32.30 \\
    & UGC & 0.8G$\sim$0.9G (\textcolor{red}{22$\times$}) & 1.5M (\textcolor{red}{36$\times$}) & \textbf{37.80} \textbf{(-48.62)} & \textbf{31.00} \textbf{(-16.46)} & \textbf{27.15} \textbf{(-11.36)} & --- \\
    & OMGD \cite{ren2021online} & 1.22G (15$\times$) & 3.4M (16$\times$) & 43.12 & 34.80 & 28.06 & 25.00 \\
    & UGC & 1.2G$\sim$1.3G (\textcolor{red}{15$\times$}) & 2.0M (\textcolor{red}{27$\times$}) & \textbf{34.80} \textbf{(-51.62)} & \textbf{27.90} \textbf{(-19.56)} & \textbf{23.25} \textbf{(-15.26)} & --- \\
    \bottomrule
    \end{tabular}
    \renewcommand{\arraystretch}{1.}
    \vspace{-5pt}
    \label{table:result1} 
\end{table*}

%% file: tables/ablation_1.tex
\begin{table}[ht]
    \centering
    \footnotesize
    \caption{Ablation studies on Semi-Supervised Learning (SSL).}
    \vspace{5pt}
    \begin{tabular}{c|l|m{0.8cm}<{\centering}m{0.8cm}<{\centering}}
    \toprule
    \multirow{2}{*}{Dataset} & \multirow{2}{*}{Method} & \multicolumn{2}{c}{Label Constraints} \\
    \cline{3-4}
    & & 10\% & 25\%  \\
    \hline
    \multirow{4}{*}{Cityscapes} 
    & Ours w/o NAS-SSL & 44.15 & 45.93 \\
    & Ours w/o Finetune-SSL  & 43.52  & 47.07 \\
    & Ours w/o ALL-SSL  & 43.51 & 45.77 \\
    \cline{2-4}
    & Ours  & 44.46 & 48.5 \\
    \midrule
    \multirow{4}{*}{Edges$\rightarrow$Shoes} 
    & Ours w/o NAS-SSL & 38.16 & 34.42 \\
    & Ours w/o Finetune-SSL  & 37.30  & 30.1 \\
    & Ours w/o ALL-SSL  & 40.57 & 35.37 \\
    \cline{2-4}
    & Ours  & 34.80 & 27.90  \\
    \bottomrule
    \end{tabular}
    \vspace{-5pt}
    \label{table:data efficiency} 
\end{table}

%% file: tables/ablation_2.tex
\begin{table}[ht]
    \centering
    \footnotesize
    \caption{Ablation studies on NAS.}
    \vspace{5pt}
    \begin{tabular}{c|c|c|c|c}
    \toprule
    \multirow{2}{*}{Dataset} & \multirow{2}{*}{MACs} &
    \multirow{2}{*}{Method} & \multicolumn{2}{c}{Label Constraints} \\
    \cline{4-5}
    & & & 10\% & 25\% \\
    
    \hline
    \multirow{4}{*}{Cityscapes} 
    & \multirow{2}{*}{1.2G} 
    & Ours w/o NAS & 42.04 & 44.64 \\
    & & Ours & 44.46 & 48.50 \\
    \cline{2-5}
    & \multirow{2}{*}{0.8G} 
    & Ours w/o NAS & 38.57 & 42.88 \\
    & & Ours & 44.83 & 47.80 \\
    \midrule
    \multirow{4}{*}{Edges$\rightarrow$Shoes} 
    & \multirow{2}{*}{1.2G} 
    & Ours w/o NAS & 44.4 & 34.77 \\
    & & Ours & 34.80 & 27.90 \\
    \cline{2-5}
    & \multirow{2}{*}{0.8G} 
    & Ours w/o NAS & 66.30 & 41.10 \\
    & & Ours & 37.80 & 31.00 \\
    \bottomrule
    \end{tabular}
    \vspace{-5pt}
    \label{table:model efficiency} 
\end{table}

%% file: tables/speed.tex
\begin{table}[t]
\centering
\footnotesize
\caption{Latency Speedup on Mobile Phones.}
\vspace{5pt}
\renewcommand{\arraystretch}{1.2}
\begin{tabular}{m{.75cm}<{\centering}cccm{0.85cm}<{\centering}m{0.85cm}<{\centering}}
\toprule
\multirow{2}{*}{\scriptsize{Model}} & \multirow{2}{*}{\scriptsize{Method}} & \multirow{2}{*}{MACs(G)} & \multicolumn{2}{c}{Mi 10 Latency(ms)}\\
\cline{4-5} 
 & & & CPU & GPU \\ 
\midrule
\multirow{4}{*}{\scriptsize{Pix2Pix}} 
& \scriptsize{Original\cite{pix2pix}} & 56.8 & 601 & 217.8\\ 
\cline{2-5} 
 & \scriptsize{GAN Com.\cite{GAN_compression}} & 5.42(10.5$\times$) & 177.6(3.4$\times$)&52.3(4.2$\times$) \\ 
 \cline{2-5} 
 & \multirow{2}{*}{\scriptsize{UGC}} & 1.50(38.0$\times$) & 92.3(6.5$\times$) & 34.1(6.4$\times$) \\ 
 \cline{3-5} 
 & & \textbf{0.89(63.6$\times$)}  & \textbf{60.7(9.9$\times$)} & \textbf{23.8(9.1$\times$)} \\ 
 
\midrule

\multirow{4}{*}{\scriptsize{GauGAN}}  
 & \scriptsize{Original\cite{park2019SPADE}} & 281.4  & 2366.7 & 1252.5 \\ 
 \cline{2-5} 
 & \scriptsize{GAN Com.\cite{GAN_compression}} & 31.2(9.0$\times$) & 549.1(4.3$\times$) & 227.7(5.5$\times$)  \\ 
 \cline{2-5} 
 & \multirow{2}{*}{\scriptsize{UGC}}  & 13.6(21.1$\times$)  & 332.5(7.1$\times$) & 155.8(8.0$\times$)\\ 
 \cline{3-5} 
 & & \textbf{9.5(29.6$\times$)}  & \textbf{275.6(8.6$\times$)} & \textbf{137.8(9.1$\times$)} \\
\bottomrule
\vspace{-50pt}
\end{tabular}
\renewcommand{\arraystretch}{1.}
\label{table:speed}
\end{table}

%% file: sections/5_Conclusion.tex
\section{Conclusion}
In this paper, we propose a new learning paradigm, Unified GAN Compression (UGC), with a unified optimization objective to seamlessly prompt the synergy of model-efficient and 
label-efficient learning.
 UGC sets up semi-supervised-driven network architecture search and adaptive online semi-supervised distillation stages sequentially, which formulates a heterogeneous mutual learning scheme to obtain an architecture-flexible, 
label-efficient, and performance-excellent model.
Extensive experiments demonstrate that UGC can obtain state-of-the-art lightweight models even with less than \textbf{50\%} labels. UGC provides a feasible solution for deploying the GAN model in the hardware-constrained and 
low-annotation regime.

%% file: UGC.bbl
\begin{thebibliography}{10}\itemsep=-1pt

\bibitem{WGAN}
Mart{\'i}n Arjovsky, Soumith Chintala, and L. Bottou.
\newblock Wasserstein gan.
\newblock {\em ArXiv}, abs/1701.07875, 2017.

\bibitem{berthelot2019remixmatch}
David Berthelot, Nicholas Carlini, Ekin~D Cubuk, Alex Kurakin, Kihyuk Sohn, Han
  Zhang, and Colin Raffel.
\newblock Remixmatch: Semi-supervised learning with distribution alignment and
  augmentation anchoring.
\newblock {\em arXiv preprint arXiv:1911.09785}, 2019.

\bibitem{imagegeneration}
Andrew Brock, Jeff Donahue, and Karen Simonyan.
\newblock Large scale gan training for high fidelity natural image synthesis.
\newblock {\em arXiv preprint arXiv:1809.11096}, 2018.

\bibitem{bigGANs}
A. Brock, J. Donahue, and K. Simonyan.
\newblock Large scale gan training for high fidelity natural image synthesis.
\newblock {\em ArXiv}, abs/1809.11096, 2019.

\bibitem{cao2021remix}
Jie Cao, Luanxuan Hou, Ming-Hsuan Yang, Ran He, and Zhenan Sun.
\newblock Remix: Towards image-to-image translation with limited data.
\newblock In {\em Proceedings of the IEEE/CVF Conference on Computer Vision and
  Pattern Recognition}, pages 15018--15027, 2021.

\bibitem{CartoonGAN}
Yang Chen, Yu-Kun Lai, and Yong-Jin Liu.
\newblock Cartoongan: Generative adversarial networks for photo cartoonization.
\newblock In {\em Proceedings of the IEEE Conference on Computer Vision and
  Pattern Recognition (CVPR)}, June 2018.

\bibitem{StarGAN}
Yunjey Choi, Minje Choi, Munyoung Kim, Jung-Woo Ha, Sunghun Kim, and Jaegul
  Choo.
\newblock Stargan: Unified generative adversarial networks for multi-domain
  image-to-image translation.
\newblock In {\em Proceedings of the IEEE Conference on Computer Vision and
  Pattern Recognition (CVPR)}, June 2018.

\bibitem{Cordts2016TheCD}
Marius Cordts, Mohamed Omran, Sebastian Ramos, Timo Rehfeld, M. Enzweiler,
  Rodrigo Benenson, Uwe Franke, S. Roth, and B. Schiele.
\newblock The cityscapes dataset for semantic urban scene understanding.
\newblock {\em 2016 IEEE Conference on Computer Vision and Pattern Recognition
  (CVPR)}, pages 3213--3223, 2016.

\bibitem{dinh2022hyperinverter}
Tan~M Dinh, Anh~Tuan Tran, Rang Nguyen, and Binh-Son Hua.
\newblock Hyperinverter: Improving stylegan inversion via hypernetwork.
\newblock In {\em Proceedings of the IEEE/CVF Conference on Computer Vision and
  Pattern Recognition}, pages 11389--11398, 2022.

\bibitem{AGD}
Y. Fu, Wuyang Chen, Haotao Wang, Haoran Li, Yingyan Lin, and Zhangyang Wang.
\newblock Autogan-distiller: Searching to compress generative adversarial
  networks.
\newblock In {\em ICML}, 2020.

\bibitem{AutoGAN}
X. Gong, S. Chang, Yifan Jiang, and Zhangyang Wang.
\newblock Autogan: Neural architecture search for generative adversarial
  networks.
\newblock {\em 2019 IEEE/CVF International Conference on Computer Vision
  (ICCV)}, pages 3223--3233, 2019.

\bibitem{gan}
Ian Goodfellow, Jean Pouget-Abadie, Mehdi Mirza, Bing Xu, David Warde-Farley,
  Sherjil Ozair, Aaron Courville, and Yoshua Bengio.
\newblock Generative adversarial nets.
\newblock In Z. Ghahramani, M. Welling, C. Cortes, N. Lawrence, and K.~Q.
  Weinberger, editors, {\em Advances in Neural Information Processing Systems},
  volume~27. Curran Associates, Inc., 2014.

\bibitem{grigoryev2022and}
Timofey Grigoryev, Andrey Voynov, and Artem Babenko.
\newblock When, why, and which pretrained gans are useful?
\newblock {\em arXiv preprint arXiv:2202.08937}, 2022.

\bibitem{resnet}
K. {He}, X. {Zhang}, S. {Ren}, and J. {Sun}.
\newblock Deep residual learning for image recognition.
\newblock In {\em 2016 IEEE Conference on Computer Vision and Pattern
  Recognition (CVPR)}, pages 770--778, 2016.

\bibitem{FID}
Martin Heusel, Hubert Ramsauer, Thomas Unterthiner, Bernhard Nessler, and S.
  Hochreiter.
\newblock Gans trained by a two time-scale update rule converge to a local nash
  equilibrium.
\newblock In {\em NIPS}, 2017.

\bibitem{slimmable_GAN}
L. Hou, Ze-Huan Yuan, Lei Huang, Hua-Wei Shen, Xueqi Cheng, and Changhu Wang.
\newblock Slimmable generative adversarial networks.
\newblock {\em ArXiv}, abs/2012.05660, 2020.

\bibitem{pix2pix}
P. {Isola}, J. {Zhu}, T. {Zhou}, and A.~A. {Efros}.
\newblock Image-to-image translation with conditional adversarial networks.
\newblock In {\em 2017 IEEE Conference on Computer Vision and Pattern
  Recognition (CVPR)}, pages 5967--5976, 2017.

\bibitem{CAT}
Qing Jin, Jian Ren, Oliver~J. Woodford, and Jiazhuo Wang.
\newblock Teachers do more than teach: Compressing image-to-image models.
\newblock {\em 2021 IEEE/CVF Conference on Computer Vision and Pattern
  Recognition (CVPR)}, 2021.

\bibitem{Perceptual}
J. Johnson, Alexandre Alahi, and Li Fei-Fei.
\newblock Perceptual losses for real-time style transfer and super-resolution.
\newblock In {\em ECCV}, 2016.

\bibitem{karras2020training}
Tero Karras, Miika Aittala, Janne Hellsten, Samuli Laine, Jaakko Lehtinen, and
  Timo Aila.
\newblock Training generative adversarial networks with limited data.
\newblock {\em Advances in Neural Information Processing Systems},
  33:12104--12114, 2020.

\bibitem{StyleGAN}
Tero Karras, Samuli Laine, and Timo Aila.
\newblock A style-based generator architecture for generative adversarial
  networks.
\newblock In {\em Proceedings of the IEEE/CVF Conference on Computer Vision and
  Pattern Recognition (CVPR)}, June 2019.

\bibitem{Karras2019stylegan2}
Tero Karras, Samuli Laine, Miika Aittala, Janne Hellsten, Jaakko Lehtinen, and
  Timo Aila.
\newblock Analyzing and improving the image quality of {StyleGAN}.
\newblock In {\em Proc. CVPR}, 2020.

\bibitem{GAN_compression}
Muyang Li, J. Lin, Yaoyao Ding, Zhijian Liu, Jun-Yan Zhu, and Song Han.
\newblock Gan compression: Efficient architectures for interactive conditional
  gans.
\newblock {\em 2020 IEEE/CVF Conference on Computer Vision and Pattern
  Recognition (CVPR)}, pages 5283--5293, 2020.

\bibitem{li2022efficient}
Muyang Li, Ji Lin, Chenlin Meng, Stefano Ermon, Song Han, and Jun-Yan Zhu.
\newblock Efficient spatially sparse inference for conditional gans and
  diffusion models.
\newblock {\em arXiv preprint arXiv:2211.02048}, 2022.

\bibitem{DMAD}
Shaojie Li, Mingbao Lin, Yan Wang, M. Xu, Feiyue Huang, Yongjian Wu, Ling Shao,
  and Rongrong Ji.
\newblock Learning efficient gans using differentiable masks and co-attention
  distillation.
\newblock {\em ArXiv}, abs/2011.08382, 2020.

\bibitem{li2021revisiting}
Shaojie Li, Jie Wu, Xuefeng Xiao, Fei Chao, Xudong Mao, and Rongrong Ji.
\newblock Revisiting discriminator in gan compression: A
  generator-discriminator cooperative compression scheme.
\newblock {\em Advances in Neural Information Processing Systems},
  34:28560--28572, 2021.

\bibitem{li2022fakeclr}
Ziqiang Li, Chaoyue Wang, Heliang Zheng, Jing Zhang, and Bin Li.
\newblock Fakeclr: Exploring contrastive learning for solving latent
  discontinuity in data-efficient gans.
\newblock In {\em European Conference on Computer Vision}, pages 598--615.
  Springer, 2022.

\bibitem{SNGAN}
Takeru Miyato, T. Kataoka, Masanori Koyama, and Y. Yoshida.
\newblock Spectral normalization for generative adversarial networks.
\newblock {\em ArXiv}, abs/1802.05957, 2018.

\bibitem{ojha2021few}
Utkarsh Ojha, Yijun Li, Jingwan Lu, Alexei~A Efros, Yong~Jae Lee, Eli
  Shechtman, and Richard Zhang.
\newblock Few-shot image generation via cross-domain correspondence.
\newblock In {\em Proceedings of the IEEE/CVF Conference on Computer Vision and
  Pattern Recognition}, pages 10743--10752, 2021.

\bibitem{GauGAN}
Taesung Park, Ming-Yu Liu, Ting-Chun Wang, and Jun-Yan Zhu.
\newblock Semantic image synthesis with spatially-adaptive normalization.
\newblock In {\em Proceedings of the IEEE/CVF Conference on Computer Vision and
  Pattern Recognition (CVPR)}, June 2019.

\bibitem{park2019SPADE}
Taesung Park, Ming-Yu Liu, Ting-Chun Wang, and Jun-Yan Zhu.
\newblock Semantic image synthesis with spatially-adaptive normalization.
\newblock In {\em Proceedings of the IEEE Conference on Computer Vision and
  Pattern Recognition}, 2019.

\bibitem{DCGAN}
A. Radford, Luke Metz, and Soumith Chintala.
\newblock Unsupervised representation learning with deep convolutional
  generative adversarial networks.
\newblock {\em CoRR}, abs/1511.06434, 2016.

\bibitem{DCGANs}
A. Radford, Luke Metz, and Soumith Chintala.
\newblock Unsupervised representation learning with deep convolutional
  generative adversarial networks.
\newblock {\em CoRR}, abs/1511.06434, 2016.

\bibitem{Reed2016GenerativeAT}
Scott~E. Reed, Zeynep Akata, Xinchen Yan, L. Logeswaran, B. Schiele, and H.
  Lee.
\newblock Generative adversarial text to image synthesis.
\newblock In {\em ICML}, 2016.

\bibitem{ren2021online}
Yuxi Ren, Jie Wu, Xuefeng Xiao, and Jianchao Yang.
\newblock Online multi-granularity distillation for gan compression.
\newblock In {\em Proceedings of the IEEE/CVF International Conference on
  Computer Vision}, pages 6793--6803, 2021.

\bibitem{unet}
O. Ronneberger, P. Fischer, and T. Brox.
\newblock U-net: Convolutional networks for biomedical image segmentation.
\newblock {\em ArXiv}, abs/1505.04597, 2015.

\bibitem{TV_Loss}
L. Rudin, S. Osher, and E. Fatemi.
\newblock Nonlinear total variation based noise removal algorithms.
\newblock {\em Physica D: Nonlinear Phenomena}, 60:259--268, 1992.

\bibitem{co_evolution}
H. Shu, Yunhe Wang, Xu Jia, Kai Han, H. Chen, Chunjing Xu, Q. Tian, and Chang
  Xu.
\newblock Co-evolutionary compression for unpaired image translation.
\newblock {\em 2019 IEEE/CVF International Conference on Computer Vision
  (ICCV)}, pages 3234--3243, 2019.

\bibitem{stylegan_v}
Ivan Skorokhodov, Sergey Tulyakov, and Mohamed Elhoseiny.
\newblock Stylegan-v: A continuous video generator with the price, image
  quality and perks of stylegan2, 2021.

\bibitem{sohn2020fixmatch}
Kihyuk Sohn, David Berthelot, Nicholas Carlini, Zizhao Zhang, Han Zhang,
  Colin~A Raffel, Ekin~Dogus Cubuk, Alexey Kurakin, and Chun-Liang Li.
\newblock Fixmatch: Simplifying semi-supervised learning with consistency and
  confidence.
\newblock {\em Advances in neural information processing systems}, 33:596--608,
  2020.

\bibitem{Tang_2019_CVPR}
Hao Tang, Dan Xu, Nicu Sebe, Yanzhi Wang, Jason~J. Corso, and Yan Yan.
\newblock Multi-channel attention selection gan with cascaded semantic guidance
  for cross-view image translation.
\newblock In {\em Proceedings of the IEEE/CVF Conference on Computer Vision and
  Pattern Recognition (CVPR)}, June 2019.

\bibitem{tseng2021regularizing}
Hung-Yu Tseng, Lu Jiang, Ce Liu, Ming-Hsuan Yang, and Weilong Yang.
\newblock Regularizing generative adversarial networks under limited data.
\newblock In {\em Proceedings of the IEEE/CVF Conference on Computer Vision and
  Pattern Recognition}, pages 7921--7931, 2021.

\bibitem{wang2021attentivenas}
Dilin Wang, Meng Li, Chengyue Gong, and Vikas Chandra.
\newblock Attentivenas: Improving neural architecture search via attentive
  sampling.
\newblock In {\em Proceedings of the IEEE/CVF Conference on Computer Vision and
  Pattern Recognition}, pages 6418--6427, 2021.

\bibitem{wang2018pix2pixHD}
Ting-Chun Wang, Ming-Yu Liu, Jun-Yan Zhu, Andrew Tao, Jan Kautz, and Bryan
  Catanzaro.
\newblock High-resolution image synthesis and semantic manipulation with
  conditional gans.
\newblock In {\em Proceedings of the IEEE Conference on Computer Vision and
  Pattern Recognition}, 2018.

\bibitem{wang2020semi}
Yaxing Wang, Salman Khan, Abel Gonzalez-Garcia, Joost van~de Weijer, and
  Fahad~Shahbaz Khan.
\newblock Semi-supervised learning for few-shot image-to-image translation.
\newblock In {\em Proceedings of the IEEE/CVF Conference on Computer Vision and
  Pattern Recognition}, pages 4453--4462, 2020.

\bibitem{SSIM}
Zhou Wang, A. Bovik, H.~R. Sheikh, and Eero~P. Simoncelli.
\newblock Image quality assessment: from error visibility to structural
  similarity.
\newblock {\em IEEE Transactions on Image Processing}, 13:600--612, 2004.

\bibitem{yang2021data}
Ceyuan Yang, Yujun Shen, Yinghao Xu, and Bolei Zhou.
\newblock Data-efficient instance generation from instance discrimination.
\newblock {\em Advances in Neural Information Processing Systems},
  34:9378--9390, 2021.

\bibitem{Yu2014FineGrainedVC}
A. Yu and K. Grauman.
\newblock Fine-grained visual comparisons with local learning.
\newblock {\em 2014 IEEE Conference on Computer Vision and Pattern
  Recognition}, pages 192--199, 2014.

\bibitem{yu2020bignas}
Jiahui Yu, Pengchong Jin, Hanxiao Liu, Gabriel Bender, Pieter-Jan Kindermans,
  Mingxing Tan, Thomas Huang, Xiaodan Song, Ruoming Pang, and Quoc Le.
\newblock Bignas: Scaling up neural architecture search with big single-stage
  models.
\newblock In {\em European Conference on Computer Vision}, pages 702--717.
  Springer, 2020.

\bibitem{Self-Attention-GAN}
Han Zhang, Ian~J. Goodfellow, Dimitris~N. Metaxas, and Augustus Odena.
\newblock Self-attention generative adversarial networks.
\newblock In {\em ICML}, 2019.

\bibitem{zhang2022wavelet}
Linfeng Zhang, Xin Chen, Xiaobing Tu, Pengfei Wan, Ning Xu, and Kaisheng Ma.
\newblock Wavelet knowledge distillation: Towards efficient image-to-image
  translation.
\newblock In {\em Proceedings of the IEEE/CVF Conference on Computer Vision and
  Pattern Recognition}, pages 12464--12474, 2022.

\bibitem{zhao2020differentiable}
Shengyu Zhao, Zhijian Liu, Ji Lin, Jun-Yan Zhu, and Song Han.
\newblock Differentiable augmentation for data-efficient gan training.
\newblock {\em Advances in Neural Information Processing Systems},
  33:7559--7570, 2020.

\bibitem{zhao2022closer}
Yunqing Zhao, Henghui Ding, Houjing Huang, and Ngai-Man Cheung.
\newblock A closer look at few-shot image generation.
\newblock In {\em Proceedings of the IEEE/CVF Conference on Computer Vision and
  Pattern Recognition}, pages 9140--9150, 2022.

\bibitem{zhao2020image}
Zhengli Zhao, Zizhao Zhang, Ting Chen, Sameer Singh, and Han Zhang.
\newblock Image augmentations for gan training.
\newblock {\em arXiv preprint arXiv:2006.02595}, 2020.

\bibitem{cyclegan}
J. {Zhu}, T. {Park}, P. {Isola}, and A.~A. {Efros}.
\newblock Unpaired image-to-image translation using cycle-consistent
  adversarial networks.
\newblock In {\em 2017 IEEE International Conference on Computer Vision
  (ICCV)}, pages 2242--2251, 2017.

\end{thebibliography}
